**Fairness Hub Technical Briefs**

# Definition and Detection of Distribution Shift

Nicolas Acevedo, Carmen Cortez, Chris Brooks, Rene Kizilcec, Renzhe Yu

Distribution shift is a common situation in machine learning tasks, where the data used for training a model is different from the data the model is applied to in the real world. This issue arises across multiple technical settings- from standard prediction tasks, to time-series forecasting, and to more recent applications of large language models (LLMs). This mismatch can lead to performance reductions, and can be related to a multiplicity of factors: sampling issues and non-representative data, changes in the environment or policies, or the emergence of previously unseen scenarios. This brief focuses on the definition and detection of distribution shifts in educational settings. We focus on standard prediction problems, where the task is to learn a model with parameters θ takes in a series of input (predictors) $X = (x_1, x_2, \cdots, x_m)$ and produces an output $\hat{Y} = f(X; \theta)$.

## For LEVI teams: Why distribution shifts?

The LEVI teams use a very wide range of AI/ML models to improve math achievement in middle schools serving vulnerable students. These models are trained on data spanning multiple different contexts (schools, interventions, moments in time, etc.), and inferences made in one context may not be generalizable to other settings. For example,

- Schools may be fundamentally different. For example, two given schools may be located in different districts or states, facing different levels of funding or oversight. When a model trained in a given school is transferred to a school in another location, the model will not account for this change in institutional characteristics, leading to decreased accuracy.
- Interventions may change the relationship between student characteristics and outcomes. For example, a given school may introduce a certain teaching method, or provide different incentives to teachers, that affect student outcomes. As these interventions may not be introduced or may not replicate exactly in another school, the model will face reduced performance when transferred.
- Distribution shifts also arise over time. If teacher characteristics, student composition, or other characteristics change within a school over time, a model trained on earlier data will also see its performance decrease at a later time.

To address this common situation, we offer a categorization of distribution shifts, along with a set of strategies to detect this issue, to create a common ground for LEVI teams to reflect on potential challenges and responses and share strategies to address them.

# Types of distribution shift

In standard machine learning tasks, when we train a model on a dataset (source) and apply it to another (target), we assume that the data distribution in the target domain is similar, if not identical, to the source domain. In other words, the joint distribution of features and labels, $p(X, y)$, remains somewhat constant across settings. Distribution shift describes a setting where this assumption does not hold, which is true for most real-world scenarios. How this occurs is usually described in one of three settings, depending on whether it affects the features, labels, or both:

1. **Covariate shift:** a change in $p(X)$, the marginal distribution of the features, with the conditional distribution $p(y|X)$ remaining stable. In this case, the feature distribution changes, while the label distribution does not. This includes subproblems such as domain shift and subpopulation shift, or sampling bias and representation bias. For example, student characteristics may change year to year, while the test score distribution remains stable.
2. **Label shift:** a change in $p(y)$, the marginal distribution of the labels, so the frequency of the labels differs between training and testing data, but the conditional distribution $p(X|y)$ remains stable. This can occur due to sampling issues, where the training data is not representative of the population, or due to changes in measurement of the label, such as redesigned tests leading to changes in scores distributions.
3. **Concept shift:** a change in $p(y|X)$, the conditional distribution of the label given the feature. This can occur when there are changes in mechanism or underlying model linking labels and features. For example, the introduction of an after-school tutoring program may change the relationship between teacher characteristics and children's test scores, degrading the predictive power of teacher characteristics.

The first figure below illustrates covariate or target shift, while the second figure shows concept shift.

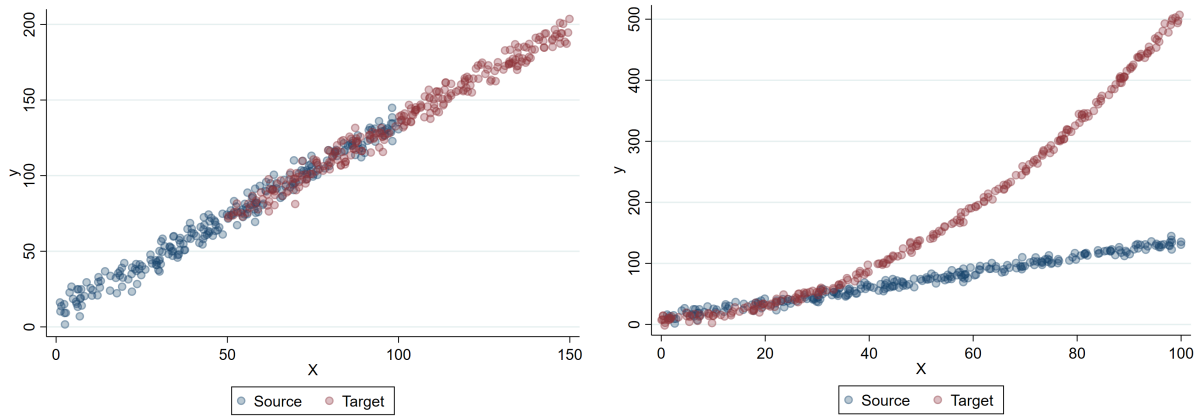

Furthermore, distribution shift can occur in static or dynamic settings:

- In static scenarios, distribution shift occurs when the distribution $p(X, y)$ is different between training and testing data, but is otherwise constant over time.
- In dynamic scenarios, distribution shift occurs when the distribution $p(X, y)$ is not constant over time, leading to differences between training and testing data.

|  | **Covariate** | **Label** | **Concept** |
|---|---|---|---|
| **Static** | The distribution of features, $p(X)$, changes between training and test data. | The distribution of the target, $p(y)$, changes between training and test data. | The conditional expectation of y, $p(y|X)$, changes between the training and test data. |
| **Dynamic** | The distribution of features, $p(X)$, changes over time, leading to continuous change in the test data. | The distribution of labels, $p(y)$, changes over time, leading to continuous change in the test data. | The conditional expectation of y, $p(y|X)$, changes over time, leading to continuous change in the test data. |

## Covariate shift

Covariate shift, as previously described, refers to changes in the distribution of the features, $p(X)$. Covariate shift may be broadly categorized as one of two types of shifts:

- Domain shift: the distribution of features between training and test data do not overlap - for example, testing a model trained using elementary school data on high school students.
- Subpopulation shift: the support of the distribution does not change, but the distribution over subgroups or subpopulations does - for example, different schools will have different socio-economic compositions.

### Detecting covariate shift

The detection of distribution shift is closely related to statistical work on measuring the distance between distributions, either via formal statistical testing, less formal visualization-based methods, or nonparametric methods. Visualization methods are the first stage in detecting covariate shifts, but any judgments are necessarily qualitative by definition. Then, the proposed measures of distance between distributions provide quantitative measures of covariate shift. Finally, statistical tests provide formal confirmation of whether two distributions are different, but require imposing additional assumptions and modeling choices.

**Visualization methods**

- **Principal components analysis** - by reducing the dimensionality of features of the training and test data, it is possible to visualize how the samples differ across their principal components by data source, as in Rabanser et al. (2019).
- **Cumulative distribution function plots** - by plotting directly the cumulative distribution function of features in the train and test data, it is possible to identify visually how different features change, as well as identify which subgroups are most affected as in Rabanser et al. (2019).

**Measures of distance between distributions**

- **Kullback-Leibler divergence** - a comparison of the difference between two probability distributions, (specifically, the expectation of the log-difference of two distributions). This divergence is calculated as:

$$D_{KL}(P||Q) = \sum P(X) \log(\frac{P(X)}{Q(X)}),$$

where $P(X)$ and $Q(X)$ are the probability density functions of the covariates in the training and testing sample, respectively. This measure has a range between zero and infinity - zero meaning that two distributions are exactly the same, with increasing values implying greater dissimilarity and infinity meaning that some events happen in a distribution but are outside the support of the other. However, this number is not a true measure of distance, as it is not symmetric - the K-L divergence between $P$ and $Q$ is not the same as the divergence between $Q$ and $P$ (Tian et al.,2021).

- **Maximum mean discrepancy:** as in Gretton et al. (2012), this is a true measure of the distance between distributions, this metric does not compare distributions directly. Instead, it projects two distributions using a kernel (uniform, Gaussian, etc.) and computes the Euclidean distance (square root of the sum of squares) between the means of the transformed distributions. Formally, the MMD is calculated as

$$MMD_k(P, Q) = ||\mu_P - \mu_Q||,$$

where $\mu_P$ and $\mu_Q$ are the means of the kernel transformations of the probability density functions of the covariates in the training and testing sample, respectively. A zero MMD implies that two distributions are identical, while larger values imply a larger shift.

- **Least-squares density difference** - estimating a function instead of a single measure as in Sugiyama et al. (2013) or in Bu et al. (2016), a least-squares density difference estimator generates a function that minimizes the sum of squared differences between two distributions. Formally, this procedure aims to estimate

$$\underset{g}{argmin} \int ((g(x) - (p(x) - q(x)))^2 dx$$

Where $g(x)$ is the objective function to be estimated and $p(x)$, $q(x)$ are the distributions of covariates in the training and testing data. If the objective function $g(x)$ is assumed to be a linear combination of Gaussian kernel-transformed data, there exists a closed form solution for this function. The strengths of this method lie in that it provides a function, instead of a single measure, providing additional information on which parts of the population may be most affected by covariate shift. However, this method requires additional assumptions, such as kernel choice or choice of regularization for the objective function, that may affect the robustness of its results.

**Statistical tests**

- **Kolmogorov-Smirnov test** - a formal statistical test based on the distance between the observed and specified cumulative distribution function, most recently discussed by Breck et al. (2019). This distance is a test statistic with a known distribution, which can be used for classical hypothesis testing, with the null hypothesis being that the specified and the observed data come from the same distribution. However, this test is restricted to univariate distributions.
- **Classifier based methods -** by setting up a classifier and testing whether it correctly predicts whether a given observation comes from the training or testing data based on the observed features, it is possible to identify covariate shift directly. This identification, as in Liu et al. (2020), is carried out by a direct statistical test based on the classifier's accuracy. However, this approach requires additional assumptions on the structure of the classifier, such as prediction thresholds.

# Label shift

In contrast to covariate shift, label shift (also known as target shift) makes the opposite assumption, where $p(X, y)$ changes, but the conditional distribution $p(X|y)$ is fixed. This situation can be found in educational settings when outcomes distributions differ across school districts due to different grading standards, but the relationship between student features and achievement, $p(X|y)$, remains constant across settings. Furthermore, the techniques described for detecting covariate shift, such as visualization, distribution distance measures, or formal statistical tests, can also be used to identify label shift. For example, it is possible to plot the cumulative distribution function plot of the label, compute any of the described distribution distance metrics, or carry out a Kolmogorov-Smirnov or a classifier test on the label distribution between training and testing data.

# Concept shift

Concept shift arises when the conditional distribution of the label given the feature, $p(y|X)$, changes either between source and target data, or over time (Lu et al., 2018). Static concept shift is linked to sampling issues that lead to the conditional relationship being different between training and testing settings. However, concept shift arises most often in dynamic settings, as the conditional relationship changes over time, due to changes in the conditions that students face, unexpected events, policy changes, or changes in student behavior.

# Detecting concept shift

- **Error rate based detection**: This approach is also known as statistical process control, and is based on monitoring the error rate of a classifier given new data and the error rate of the same classifier using a baseline sample, by assuming that error rates are a Bernoulli random variable with a binomial distribution.
  In the simplest setting, $p_t$ is the instantaneous prediction error rate, with a standard deviation given by $s_t = \sqrt{\frac{p_t(1-p_t)}{i}}$. The model defines a warning state when $p_t + s_t \geq p_{min} + 2s_{min}$, and a drift when $p_t + s_t \geq p_{min} + 3s_{min}$ - i.e., when error rates are two or three standard deviations above their historical minima (Gama et al., 2004). More recent implementations of this approach hinge on the choice of sample to estimate error rates, the calculation of optimal window widths, and the choice fixed versus moving-average sizes of the baseline sample.

- **Multiple hypothesis testing detection**: adapts error based detection to multiple measures of error rate or drift, such as AUC, F1 scores, etc., adapting the procedure to the strengths and weaknesses of each error rate measure, as in Wang and Abraham (2015). This multiple hypothesis testing can be conducted either in parallel or serially, and can be extended to test changes in features instead of metrics.
- **Distribution drift detection**: this approach is only applicable in dynamic settings, where concept shift happens over time. This approach calculates distribution distance measures between features in the original data, used as a training sample, and the updated data with additional time periods as the testing sample. In this setting, visualization methods, calculated measures of distance between distributions such as the Kullback-Leibler divergence, as in Dasu et al. (2006) and statistical tests are applicable.

# How to diagnose different types of shifts in practice

Recent research has focused on quantifying the relative importance of covariate shifts and concept shifts in performance loss, as in Cai (2023). This can be achieved by decomposing the performance degradation by quantifying the loss in the common support of the training and testing distributions, $P$ and $Q$, and the loss outside the common support for each distribution.

More specifically, allow $l(f(X), y)$ to be any loss function of a model trained with data $X$ on outcome $y$. This loss can be calculated for models trained on both the training and testing distributions, $P$ and $Q$. The relevant object is the conditional expectation of this loss function over both distributions, called $R$, also interpreted as the conditional risk.

$$R_P(X) = E_P[l(f(X), y)|X = x]$$
$$R_Q(X) = E_Q[l(f(X), y)|X = x]$$

To identify concept shift, we compare these risks across the same marginal distribution of $X$ - if the distribution of $X$ is kept constant, the difference between $R_P(X)$ and $R_Q(X)$ is solely due to changes in $Y|X$. However, to keep the distribution of $X$ constant across the training and testing distribution, it is necessary to create a distribution S that belongs to the common support of $P$ and $Q$. Based on this distribution $S$, it is possible to quantify the loss in performance due to $Y|X$ by comparing $R_P(X)$ and $R_Q(X)$ by calculating them over the shared distribution $S$.

More generally, it is possible to decompose performance loss into the component caused by covariate shift, the component caused by concept shift, and a component caused by out of sample prediction. Formally,

$$E_Q[l(f(X), y)] - E_P[l(f(X), y)] = (E_S[R_P(X)] - E_P[R_P(X)]) + E_S[R_Q(X) - R_P(X)] \\ + (E_Q[R_Q(X)] - E_S[R_Q(X)])$$

where the first term quantifies covariate shift, the second term measures concept shift, and the third term measures performance loss due to previously unseen examples in the testing distribution. In practice, the first and the third term are qualitatively similar for model diagnostics.